\documentclass[runningheads]{llncs}

\usepackage{eccv}

\usepackage{eccvabbrv}

\usepackage{graphicx}
\usepackage{booktabs}

\usepackage[accsupp]{axessibility}  
\usepackage[table]{xcolor}
\usepackage{colortbl}
\newlength{\subfiga}\newlength{\subfigb}

\usepackage{hyperref}

\usepackage{orcidlink}

\begin{document}

\title{Fourier Splatting: Generalized Fourier encoded primitives for scalable radiance fields} 


\titlerunning{Fourier Splatting}

\author{
Mihnea-Bogdan Jurca\inst{1,2}\thanks{Equal contributions.}\and
Bert Van hauwermeiren\inst{1}\textsuperscript{*} \and \\
Adrian Munteanu\inst{1}}

\authorrunning{Jurca et al.}

\institute{Vrije Universiteit Brussel, Brussels, Belgium \and
Technical University of Cluj-Napoca, Cluj-Napoca, Romania
\email{\{Mihnea-Bogdan.Jurca, Bert.Karel.Van.hauwermeiren, Adrian.Munteanu\}@vub.be}}

\maketitle

\begin{figure*}[t]
  \centering
  \includegraphics[width=\textwidth]{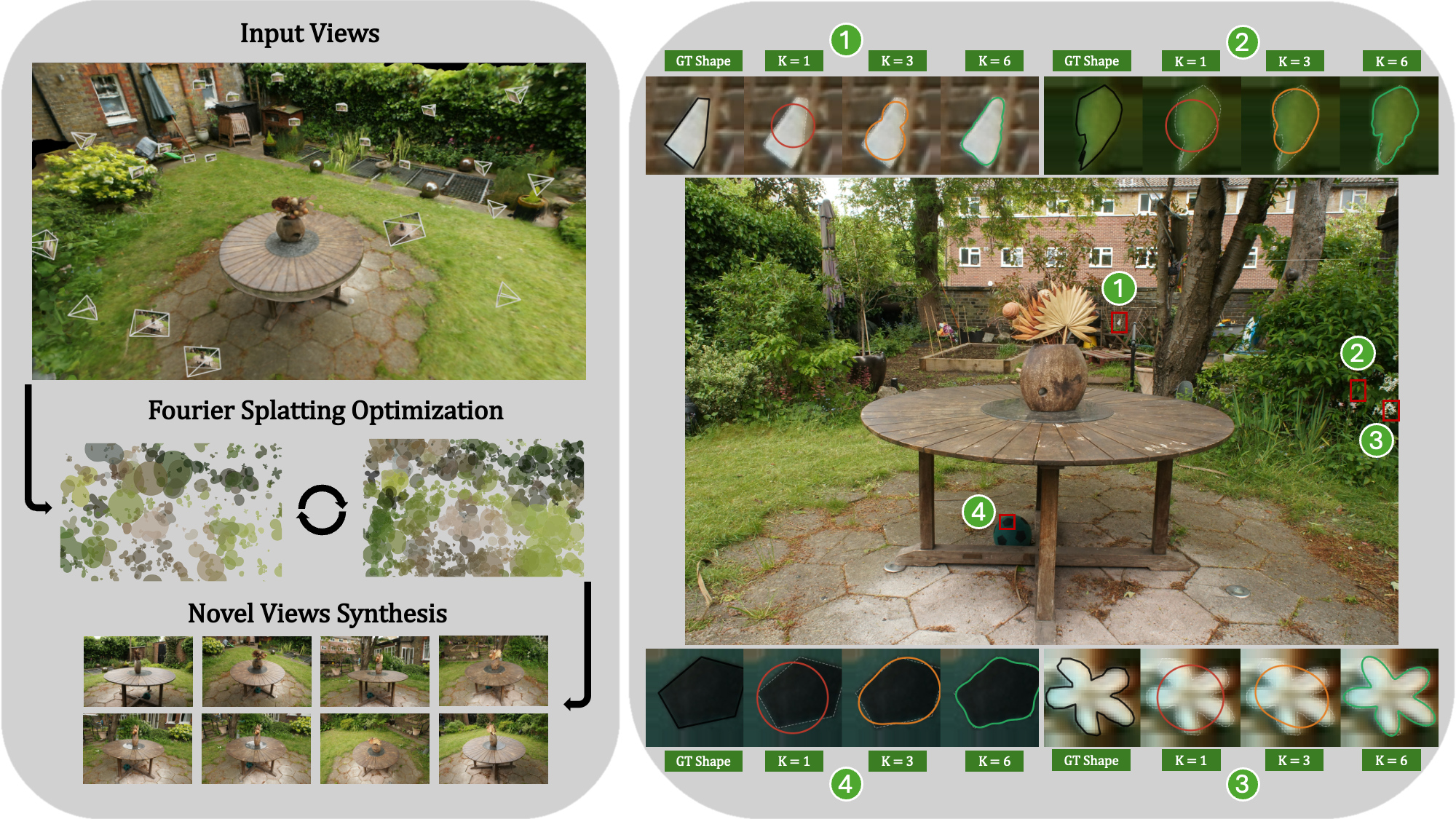}
  \caption{Overview of Fourier Splatting using generalized Fourier encoded primitives. Varying the number K of retained frequencies progressively improves the shape of the primitives. This provides an inherently scalable representation of the radiance field with progressively improved rendering quality.}
  \label{fig:pipeline}
\end{figure*}

\begin{abstract}
  Novel view synthesis has recently been revolutionized by 3D Gaussian Splatting (3DGS), which enables real-time rendering through explicit primitive rasterization. However, existing methods tie visual fidelity strictly to the number of primitives: quality downscaling is achieved only through pruning primitives. We propose the first inherently scalable primitive for radiance field rendering. {\bf Fourier Splatting} employs scalable primitives with arbitrary closed shapes obtained by parameterizing planar surfels with Fourier encoded descriptors. This formulation allows a single trained model to be rendered at varying levels of detail simply by truncating Fourier coefficients at runtime. To facilitate stable optimization, we employ a straight-through estimator for gradient extension beyond the primitive boundary, and introduce HYDRA, a densification strategy that decomposes complex primitives into simpler constituents within the MCMC framework.
  Our method achieves state-of-the-art rendering quality among planar-primitive frameworks and comparable perceptual metrics compared to leading volumetric representations on standard benchmarks, providing a versatile solution for bandwidth-constrained high-fidelity rendering.
\end{abstract}

\section{Introduction}
\label{sec:intro}

Photorealistic novel view synthesis has emerged as a cornerstone of
computer vision, underpinning downstream tasks from content
generation~\cite{tang_dreamgaussian_2024,yi_gaussiandreamer_2024 } to semantic scene
understanding~\cite{qin_langsplat_2024,ye_gaussian_2025, jurca_rt-gs2_2024}, etc.
Neural Radiance Fields
(NeRF)~\cite{mildenhall_nerf_2021} established that continuous volumetric
representations can produce high-fidelity views, but their reliance on dense ray
marching confines rendering to offline settings. 3D Gaussian Splatting
(3DGS)~\cite{kerbl_3d_2023} offered a compelling alternative:
explicit Gaussian primitives combined with tile-based rasterization enable
real-time rendering, making splatting-based methods the preferred framework for
interactive radiance fields. Given their growing impact across applications,
scalability of these representations has become one of the most pressing open
problems.\looseness=-1

Recent efforts to refine radiance field primitives have moved beyond the standard ellipsoidal 3D Gaussian, seeking representations that better capture complex geometries and sharp boundaries. These include volumetric approaches~\cite{hamdi_ges_2024, liu_deformable_2025, held_3d_2025, zhang_quadratic_2025},  with higher quality image generation and planar approaches~\cite{huang_2d_2024, held_triangle_2025, svitov_billboard_2025} that enforce surface consistency and resolve the depth ambiguities inherent in volumetric formulations. However, while these designs improve per-primitive fidelity, they remain fundamentally static in their capacity; they rely on traditional density-driven optimization, where high-frequency details are resolved through the proliferation of more primitives rather than an expansion of the primitive’s own geometric complexity.\looseness=-1

To manage the computational and memory overhead of increasingly dense splat collections, contemporary research has turned toward Level-of-Detail (LoD) hierarchies. State-of-the-art scalable frameworks typically organize Gaussians into hierarchical layers where distant or less significant primitives are pruned or filtered based on importance metrics. While effective for reducing bitrate and improving real-time performance, these methods achieve scalability only through pruning. By focusing exclusively on reducing the primitive count.\looseness=-1

To address these limitations, we present \textbf{Fourier Splatting}, the first inherently scalable primitive
for real-time radiance field rendering. Rather than relying solely on primitive
population growth to increase fidelity, we propose a novel primitive that can approximate any arbitrary shape and is inherently scalable.
Each surfel's boundary is parameterized as a Fourier encoded descriptors,
where the coefficients control progressively finer shape detail: at the lowest
order the primitive reduces to a standard disc, and each additional frequency
component enables it to express increasingly complex closed curves within a single, compact parameterization. The design
yields a natural level-of-detail (LoD) hierarchy: the same trained model can be
rendered at any target complexity simply by truncating the Fourier coefficients, without
retraining or post-hoc compression. We achieve state-of-the-art rendering among planar methods, and achieve at least similar performance to volumetric 3D representations on standard
benchmarks. 

\noindent Our primary contributions are summarized as follows:
\begin{itemize}
    \item \textbf{Arbitrary-Shape Planar Primitives:} We introduce a novel planar primitive for splatting that leverages a Fourier-based boundary parameterization. Unlike other splats, our primitives can assume arbitrary shapes, enabling the representation of complex geometry and fine-grained details with significantly higher fidelity compared to 2D and 3D Gaussians. 

    \item \textbf{Inherent Scalability:} We present the first radiance field primitive that is inherently scalable. By truncating the set of active Fourier coefficients, the level-of-detail (LoD) can be continuously adjusted at runtime without additional training or post-processing, providing direct control over the rate-distortion trade-off.

    \item \textbf{MCMC-Compatible Densification:} We propose HYDRA, a densification strategy that enables generalizable planar primitives to operate within the MCMC framework, together with a straight-through estimator that extends gradients beyond the primitive boundary to prevent optimization stalling.

    \item \textbf{State-of-the-Art Performance:} We demonstrate that our method surpasses existing planar-primitive approaches across standard benchmarks. Furthermore, our representation achieves competitive results against leading volumetric methods, outperforming in some categories. Finally, we demonstrate that our scalable primitive exhibits more graceful qualitative degradation during downscaling compared to the state-of-the-art
\end{itemize}

\section{Related Work}

{\bf Neural scene representations.}
Neural radiance fields~\cite{mildenhall_nerf_2021} established image-based 3D
reconstruction through implicit volumetric representations. While subsequent work has significantly reduced training times and improved quality~\cite{barron_mip-nerf_2022, muller_instant_2022, barron_zip-nerf_2023}, the underlying reliance on dense per-ray volume integration remains a fundamental computational bottleneck. To circumvent this, 3D Gaussian Splatting~(3DGS)~\cite{kerbl_3d_2023} introduced an explicit representation composed of millions of anisotropic Gaussians. By replacing volumetric sampling with highly optimized tile-based rasterization, 3DGS achieves real-time frame rates with state-of-the-art visual fidelity. Recent advancements have further refined the 3DGS framework by optimizing densification through MCMC sampling~\cite{kheradmand_3d_2024}, absolute gradient accumulation~\cite{ye_absgs_2024}, or by introducing structured spatial organization~\cite{lu_scaffold-gs_2024,ren_octree-gs_2025}. All these approaches use the same smooth elliptical Gaussian primitive.

{\bf Alternative primitive representations.}
An emerging branch of research proposes to move beyond the standard Gaussian kernel as the building block for splatting. GES~\cite{hamdi_ges_2024} generalizes the radial
falloff to an exponential family with a learnable shape parameter, enabling
sharper profiles. Deformable Beta Splatting~\cite{liu_deformable_2025} replaces
Gaussians with compact-support Beta kernels whose shape parameters control
the falloff from uniform to peaked. 3D-HGS~\cite{li_3d-hgs_2025} splits each Gaussian
along a plane to create asymmetric half-kernels for discontinuities.
DRK~\cite{huang_deformable_2025} parameterizes the radial profile with learnable basis
functions whose scale and sharpness vary per angular direction, producing
star-shaped footprints. Beyond kernel modifications, 3D Convex
Splatting~\cite{held_3d_2025} replaces Gaussians with smooth convex
bodies that can represent hard edges and flat surfaces, Quadratic Gaussian
Splatting~\cite{zhang_quadratic_2025} deforms a surfel from a flat disc to a
paraboloid. While these methods significantly expand the design space of the splatting unit, they share a common constraint: the expressiveness of each primitive is determined by its architectural design and remains static once trained.

{\bf Surface primitives.}
Another line of research is surface-based splatting methods, which better aligns primitives with underlying scene geometry to facilitate high-fidelity reconstruction. 2D Gaussian
Splatting~(2DGS)~\cite{huang_2d_2024} collapses 3D Gaussians to oriented planar
surfels. Gaussian Surfels~\cite{dai_high-quality_2024} similarly
flatten ellipsoids to discs and enforce normal--depth consistency through
monocular priors. Triangle Splatting~\cite{held_triangle_2025} departs
from smooth kernels entirely, rendering optimizable triangles as differentiable
splats with hard boundaries. BBSplat~\cite{svitov_billboard_2025} equips each
billboard plane with a learned texture and alpha mask, encoding spatially-varying
appearance within a single flat primitive. 


 {\bf Scalable representations.}
Orthogonal to primitive design, a separate line of work focuses on managing the
complexity of Gaussian-based scenes. Octree-gs~\cite{ren_octree-gs_2025} and LODGE~\cite{kulhanek_lodge_2025} maintain different sets of LoD Gaussians, pruning and blending Gaussians based on rendering distance. LOD-GS~\cite{shen_lod-gs_2025} learns  triangle soup representation with 3D gaussians embedded in them, scalability is achieved through downsampling the gaussians in each triangle. While adaptive densification strategies~\cite{kerbl_3d_2023,kheradmand_3d_2024,ye_absgs_2024}
control how many primitives are spawned during optimization. In all cases, the
axis of scalability is the number of primitives: representations are made
smaller by removing Gaussians and larger by adding them, while each individual
primitive remains unchanged. Our Fourier Splatting approach differs fundamentally by embedding scalability within the primitive itself; rather than discarding entire spatial regions or points, we reduce the precision of the primitives themselves.


\section{Preliminaries}

\subsection{3D Gaussian Splatting}

3DGS~\cite{kerbl_3d_2023} represents a scene as a collection of
anisotropic Gaussian primitives. Each primitive is parameterized by
a mean $\boldsymbol{\mu} \in \mathbb{R}^3$, a covariance
$\boldsymbol{\Sigma} = \mathbf{R}\mathbf{S}\mathbf{S}^\top
\mathbf{R}^\top$ factored into rotation $\mathbf{R}$ and diagonal
scale $\mathbf{S}$, an opacity $o \in (0,1)$, and spherical
harmonic coefficients for view-dependent color~$\mathbf{c}$.
The 3D covariance is projected to 2D via
$\boldsymbol{\Sigma}' = \mathbf{J}\mathbf{V}\boldsymbol{\Sigma}
\mathbf{V}^\top\mathbf{J}^\top$, where $\mathbf{V}$ is the
viewing transformation (world-to-camera) and $\mathbf{J}$ the
Jacobian of the local affine approximation of the projective
transformation. Primitives are sorted by
depth and accumulated front-to-back:

\begin{equation}
  \mathbf{C}(\mathbf{x}) = \sum_{i=1}^{N} \mathbf{c}_i \,
  \alpha_i \, T_i = \sum_{i=1}^{N} \mathbf{c}_i \,
  \alpha_i \prod_{j=1}^{i-1} (1 - \alpha_j).
  \label{eq:compositing}
\end{equation}

where $\alpha_i$ is the product of opacity and the evaluated 2D
Gaussian
$\mathcal{N}(\mathbf{x};\boldsymbol{\mu}'_i,\boldsymbol{\Sigma}'_i)$.
This enables real-time tile-based rasterization, but each primitive
is rendered as a smooth ellipsoidal kernel with infinite support and
fixed shape.

\subsection{2D Gaussian Splatting}

2DGS~\cite{huang_2d_2024} replaces volumetric Gaussians with
oriented planar surfels. Each surfel is defined by a center
$\mathbf{p}_k \in \mathbb{R}^3$, tangent vectors
$\mathbf{t}_u, \mathbf{t}_v$ with scales $s_u, s_v > 0$, and
normal $\mathbf{t}_w = \mathbf{t}_u \times \mathbf{t}_v$. A point
on the surfel is:
\begin{equation}
  \mathbf{P}(u,v) = \mathbf{p}_k + s_u\,\mathbf{t}_u\,u
  + s_v\,\mathbf{t}_v\,v = \mathbf{H}\,(u,v,1,1)^\top,
  \label{eq:surfel_param}
\end{equation}
where $\mathbf{H} \in \mathbb{R}^{4\times4}$ is the homogeneous
surfel-to-world matrix.

{\bf Ray-surfel intersection.}
Instead of projecting a covariance, 2DGS intersects each pixel ray
with the surfel plane. Let
$\mathbf{W} \in \mathbb{R}^{4\times4}$ be the world-to-screen
transformation. A pixel at $(x_p, y_p)$ defines two screen-space
planes $\mathbf{h}_x = (-1,0,0,x_p)^\top$ and
$\mathbf{h}_y = (0,-1,0,y_p)^\top$. Transforming these into the
surfel-local frame gives
$\mathbf{h}_u = (\mathbf{WH})^\top\mathbf{h}_x$ and
$\mathbf{h}_v = (\mathbf{WH})^\top\mathbf{h}_y$. The intersection
coordinates follow from solving the $2\times 2$ system:
\begin{equation}
  u_j = \frac{h_u^2 h_v^4 - h_u^4 h_v^2}
  {h_u^1 h_v^2 - h_u^2 h_v^1},\quad
  v_j = \frac{h_u^4 h_v^1 - h_u^1 h_v^4}
  {h_u^1 h_v^2 - h_u^2 h_v^1}.
  \label{eq:uv_intersection}
\end{equation}
The per-pixel opacity is then
$\alpha_i = o_i\,\exp(-(u_j^2+v_j^2)/2)$, a 2D Gaussian evaluated
in the surfel plane. Our method inherits the surfel
parameterization, ray-plane intersection, and tile-based compositing
from 2DGS, but replaces the fixed Gaussian falloff with the
Fourier boundary and power window described in
\cref{sec:primitive,sec:rendering}.

\begin{figure}[t]
  \centering
  \settoheight{\subfiga}{\includegraphics[width=0.5\linewidth]{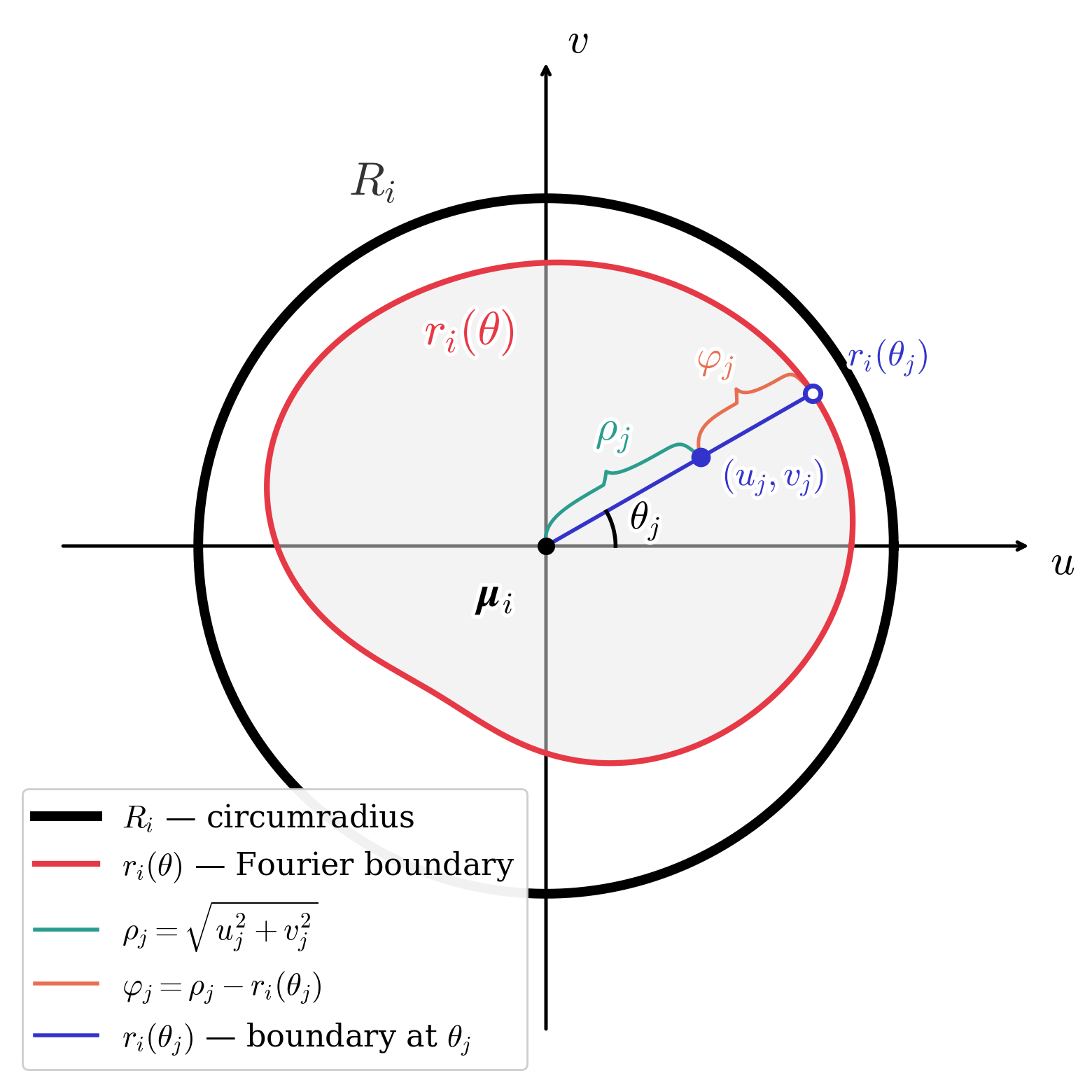}}
  \settoheight{\subfigb}{\includegraphics[width=0.5\linewidth]{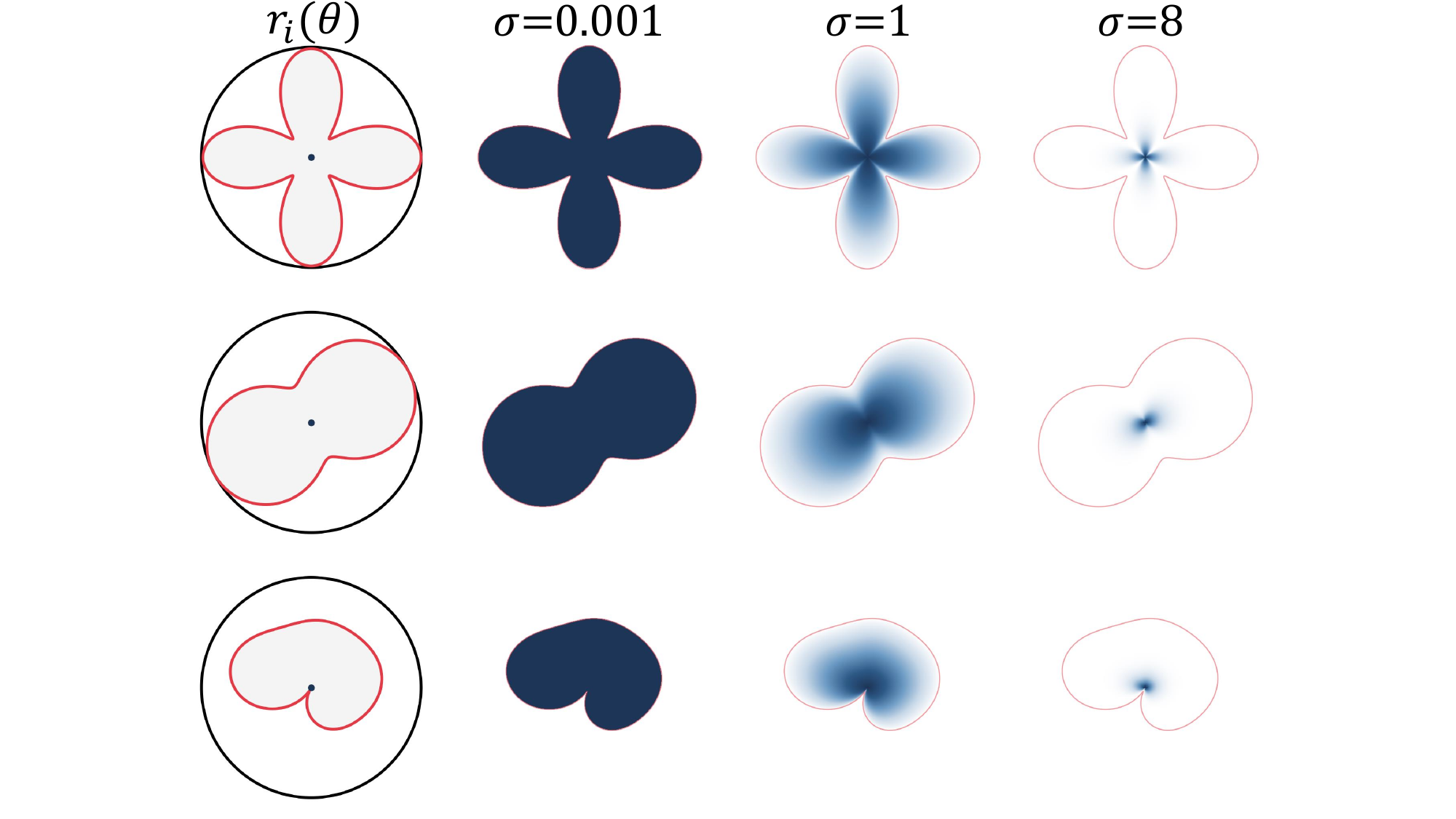}}
  \begin{subfigure}[b]{0.5\linewidth}
    \centering
    \includegraphics[width=\linewidth]{figures/fourier_primitive_clean.png}
    \caption{Primitive parameterization: Fourier boundary
    $r_i(\theta)$ within circumradius $R_i$.}
    \label{fig:primitive_param}
  \end{subfigure}%
  \hfill%
  \begin{subfigure}[b]{0.5\linewidth}
    \centering
    \raisebox{0.5\dimexpr\subfiga-\subfigb\relax}{%
      \includegraphics[width=\linewidth]{figures/Presentation3.pdf}}
    \caption{Shape diversity and effect of sharpness $\sigma_i$.}
    \label{fig:shape_diversity}
  \end{subfigure}
  \caption{Overview of the Fourier splatting primitive.}
  \label{fig:overview}
\end{figure}

\section{Method}
\label{sec:method}

Fourier Splatting builds on the surfel
formalism~\cite{zwicker_surface_2001,huang_2d_2024} but departs from the
fixed disc and ellipsoid assumptions. Given a scene represented
by a set of $N$ primitives $\mathcal{S} = \{s_i\}_{i=0}^{N-1}$, we equip each
$s_i$ with a Fourier encoded descriptors that defines an arbitrary
closed boundary within its tangent plane, making primitive expressiveness a
continuous, controllable quantity rather than a fixed design choice. We first
describe the primitive parameterization (\cref{sec:primitive}), then its
differentiable rendering (\cref{sec:rendering}), and the densification,
pruning and training strategies (\cref{sec:densification}).

\subsection{Fourier Splatting Primitive}
\label{sec:primitive}

Each primitive $s_i$ is parameterized by a center
$\mathbf{p}_i \in \mathbb{R}^3$, a rotation quaternion
$\mathbf{q}_i \in \mathbb{R}^4$ defining the local tangent frame
$(\mathbf{t}_u, \mathbf{t}_v, \mathbf{t}_w)$, an opacity
$o_i \in (0,1)$, spherical harmonic coefficients
$\mathbf{c}_i \in \mathbb{R}^{48}$ for view-dependent appearance, a
circumradius $R_i \in \mathbb{R}_{>0}$ bounding the primitive extent,
Fourier amplitudes $\{r_{i,k}\}_{k=0}^{K-1} \in \mathbb{R}^{K}$ and
phases $\{\phi_{i,k}\}_{k=0}^{K-1} \in \mathbb{R}^{K}$ defining the
boundary shape, and a sharpness control parameter $\sigma_i \in \mathbb{R}_{>0}$.

{\bf Fourier encoded boundary.}
We define the boundary of each primitive $s_i$ as a closed curve
$r_i(\theta)$ in its local tangent plane (\cref{fig:overview}a).
Given a point $(u_j, v_j)$ in the tangent plane obtained via
ray-plane intersection (\cref{eq:uv_intersection}), we compute its
polar coordinates relative to the primitive center
$\boldsymbol{\mu}_i$: the radial distance
$\rho_j = \sqrt{u_j^2 + v_j^2}$ and the angle
$\theta_j = \arctan\!\left(\frac{v_j}{u_j}\right)
$. The boundary radius at
angle $\theta_j$ is given by the modulus of a Fourier polynomial:
\begin{equation}
  r_i(\theta_j) = R_i \cdot \left|
  \sum_{k=0}^{K-1} c_{i,k} \, e^{\mathrm{i}\,k\,\theta_j} \right|,
  \label{eq:fourier_boundary}
\end{equation}
where $c_{i,k} = \bar{r}_{i,k} \, e^{\mathrm{i}\,\phi_{i,k}}$ are
complex coefficients with normalized amplitudes $\bar{r}_{i,k}$ and
phases $\phi_{i,k}$ (\textit{note}: $\mathrm{i}$ in the exponent denotes the
imaginary unit, while the subscript $i$ indexes the primitive). The
$k\!=\!0$ term acts as the DC component, setting a uniform base
radius, while each higher frequency $k$ introduces $k$-fold angular
variation. At $K\!=\!1$ only the DC term remains and the boundary
reduces to a circle of radius $R_i \bar{r}_{i,0}$, recovering the standard disc primitive. 

{\bf Amplitude normalization.}
Since each frequency $k$ contributes a radius $\bar{r}_{i,k}$ to
the boundary, the total reach of the primitive is bounded by the sum
of all amplitudes. Without normalization, these amplitudes can sum
to exceed the circumradius $R_i$, causing the boundary to overshoot
the circumscribed circle. We enforce this constraint via
squared-$\ell_1$ normalization:
\begin{equation}
  \bar{r}_{i,k} = \frac{r_{i,k}^2}
  {\sum_{k'=0}^{K-1} r_{i,k'}^2} \quad \text{such that} \quad
  \sum_{k=0}^{K-1} \bar{r}_{i,k} = 1.
  \label{eq:norm_constraint}
\end{equation}

Since each $\bar{r}_{i,k} \geq 0$ and their sum equals one, the
maximum constructive interference across all frequencies cannot
exceed $R_i$, guaranteeing $r_i(\theta) \leq R_i$ for all $\theta$. \looseness=-1


\subsection{Differentiable Rendering}
\label{sec:rendering}

{\bf Forward pass.}
We implement a tile-based rasterizer for the Fourier primitive,
adapting the power window function of Triangle
Splatting~\cite{held_triangle_2025} to our polar coordinate
formulation. For each pixel $j$ with polar coordinates
$(\rho_j, \theta_j)$ in the tangent plane of primitive $s_i$, the
signed distance to the Fourier boundary is:
\begin{equation}
  \varphi_j = \rho_j - r_i(\theta_j),
  \label{eq:signed_distance}
\end{equation}
which is negative inside and positive outside the primitive; points with
$\varphi_j \geq 0$ are discarded, providing compact support. The
per-pixel opacity is:
\begin{equation}
  \alpha_{i,j} = o_i \cdot
  \max\!\left(0,\;
    \frac{r_i(\theta_j) - \rho_j}{r_i(\theta_j)}
  \right)^{\!\sigma_i}\!.
  \label{eq:alpha}
\end{equation}
The window peaks at the primitive center and
decays to zero at the boundary, with $\sigma_i$ controlling the
decay profile. Colors are accumulated via front-to-back compositing
(\cref{eq:compositing}).


{\bf Backward pass.}
The hard clamp in \cref{eq:alpha} creates a zero-gradient region
outside the primitive boundary. For any pixel with
$\varphi_j \geq 0$, the rendered contribution is exactly zero and
the gradient with respect to the Fourier coefficients $c_{i,k}$
vanishes. The boundary therefore contracts, as most gradient signal
comes from interior pixels. To compensate, the optimization inflates
the circumradius $R_i$, scaling the entire shape uniformly rather
than letting the Fourier coefficients learn more complex geometries.

\begin{figure*}[t]
  \centering
  \includegraphics[width =\linewidth]{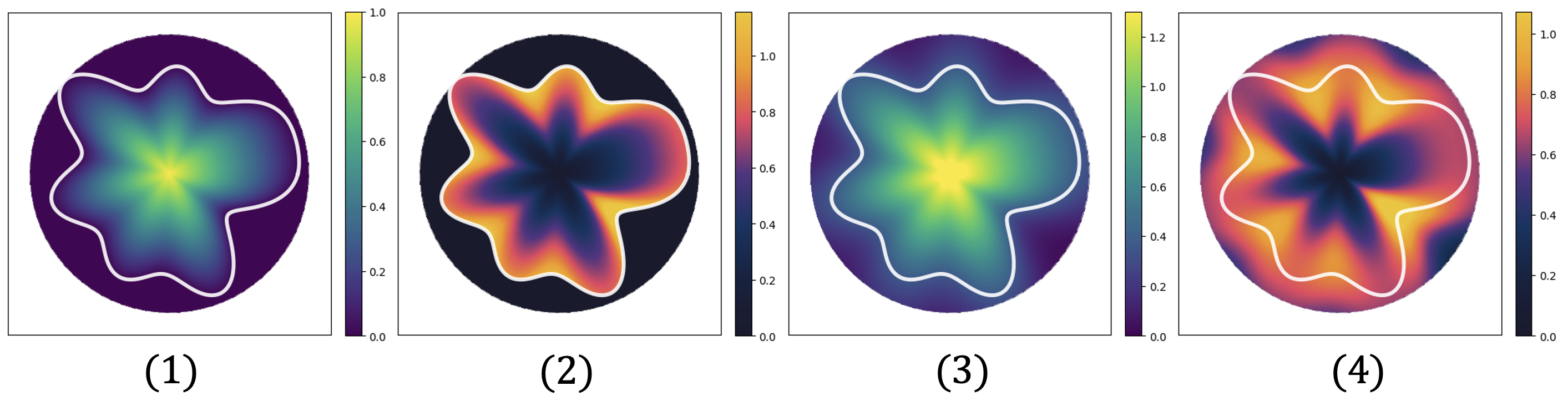}
  \caption{Straight-Through Estimator (STE) for gradient extension.
(1)~Forward rendering with the hard cutoff window (\cref{eq:alpha}).
(2)~Gradient without STE: dead zone outside the boundary.
(3)~Smooth STE surrogate (\cref{eq:ste}).
(4)~Gradient with STE: signal flows across the full primitive
extent. Solid line: $r_i(\theta)$; dashed: $R_i$.
$\beta\!=\!3$, $\gamma\!=\!0.5$, $\sigma\!=\!1$.}
  \label{fig:ste}
\end{figure*}

To address this, we draw on the straight-through estimator
(STE)~\cite{bengio_estimating_2013} and decouple the forward and
backward passes. The forward rendering is unchanged: the hard window
of \cref{eq:alpha} determines the rendered image. In the backward
pass, we replace the hard window with a smooth surrogate
$\tilde{w}$ whose gradient is non-zero on both sides of the boundary
(\cref{fig:ste}):
\begin{equation}
  \tilde{w}(x) = \operatorname{sigmoid}(\beta x)
  \cdot \left(\frac{\operatorname{softplus}(\beta x)}{\beta}
  \right)^{\!\sigma_i}
  + \;\gamma \cdot \operatorname{sigmoid}(\beta x),
  \label{eq:ste}
\end{equation}
where $x = (r_i(\theta_j) - \rho_j) / r_i(\theta_j)$ is the
unclamped normalized distance ($x > 0$ inside, $x < 0$ outside),
and $\beta$, $\gamma$ are fixed hyperparameters. The first term is a
smooth approximation of $\max(0, x)^{\sigma_i}$, matching the
forward window for interior points. The additive sigmoid term
ensures $\partial \tilde{w} / \partial x \neq 0$ even when $x < 0$,
providing gradient signal to the Fourier coefficients from pixels
just outside the boundary. This allows the boundary to both expand and contract during
optimization, letting the Fourier coefficients learn complex
shapes instead of relying solely on the circumradius $R_i$ for
spatial coverage. Since exterior pixels have zero contribution in the forward pass,
we restrict their STE gradient to the Fourier coefficients
$c_{i,k}$. Position and orientation gradients are computed
from interior pixels where the photometric loss is well-defined.
 The effect of $\beta$ and $\gamma$ on the
gradient coverage area is illustrated in \cref{fig:ste}.

\subsection{Optimization}
\label{sec:densification}

We adopt the MCMC framework of
3DGS-MCMC~\cite{kheradmand_3d_2024}, which interprets SGD updates
as Stochastic Gradient Langevin Dynamics (SGLD) and treats
primitives as samples from a target distribution
$P(\mathbf{g}) \propto \exp(-\mathcal{L})$, where $\mathbf{g}$
denotes the full primitive state. Densification is a deterministic
state transition where dead primitives are relocated to live ones
with parameters adjusted to preserve
$P(\mathbf{g}^{\mathrm{new}}) = P(\mathbf{g}^{\mathrm{old}})$.
A key ingredient of MCMC~\cite{kheradmand_3d_2024} is the SGLD noise added
to primitive positions, scaled proportionally to the covariance
$\boldsymbol{\Sigma}$ and inversely to the opacity, encouraging
spatial exploration. For planar primitives this is ill-suited: positional noise scaled
by a 3D covariance would push surfels out of their tangent plane
when they have not yet converged or lie in high volumetric-frequency
regions, and since children inherit the parent's orientation, such
perturbations violate the planar constraint and degrade
$P(\mathbf{g}^{\mathrm{new}})$.

{\bf Relocation.}
When a primitive with opacity $o$, circumradius $R$, and sharpness
$\sigma$ is replaced by $N$ co-located copies, we require the
rendered image to remain unchanged. The new opacity follows the
kernel-independent formula
$o_{\mathrm{new}} = 1 - (1 - o)^{1/N}$ \cite{kheradmand_3d_2024}. For the scale, we
equate the 2D polar integral of the parent's power window
$w(\rho) = \max(0,\,(R-\rho)/R)^{\sigma}$ with the composited
integral of $N$ children as in \cite{kheradmand_3d_2024}. Expanding the per-child transmittance
$(1-o_{\mathrm{new}}\,w)^{i-1}$ via the binomial theorem and
integrating each term $w^{k+1} = ((R-\rho)/R)^{\sigma(k+1)}$
in polar coordinates gives:
\begin{equation}
  D = \sum_{i=1}^{N}\sum_{k=0}^{i-1}
  \binom{i\!-\!1}{k}(-1)^{k}\,
  \frac{o_{\mathrm{new}}^{k+1}}
  {(\sigma(k\!+\!1)+1)(\sigma(k\!+\!1)+2)},
  \label{eq:relocation_denom}
\end{equation}
where the denominator factors replace the Gaussian integral
of~\cite{kheradmand_3d_2024}. The corrected circumradius is:
\begin{equation}
  R_{\mathrm{new}} = R\sqrt{\frac{o}
  {(\sigma+1)(\sigma+2)\cdot D}}.
  \label{eq:scale_relocation}
\end{equation}

{\bf Death criterion.}
Primitives are removed if: (i)~opacity falls below $\tau_o$,
(ii)~the maximum blending weight
$\max_j(\alpha_{i,j}\,T_{i,j})$ over all pixels $j$ across
recent views falls below $\tau_I$ (\cref{eq:compositing}), or
(iii)~primitive $s_i$ is observed in fewer than $V_{\min}$
frames \cite{held_triangle_2025}. Newly created primitives are protected by a one-epoch grace period.

{\bf Primitive relocation and addition.}
Dead primitives, identified by the death criterion, are
relocated to live ones by sampling primitives weighted by opacity or
inverse sharpness $1/\sigma_i$ in alternation due to different
local optimization convergence~\cite{held_triangle_2025}. Each
sampled primitive produces two children via the relocation formulas
of \cref{eq:relocation_denom,eq:scale_relocation} with
$N\!=\!2$: one at the sampled position, the other offset in the primitive's
tangent plane by $\eta \boldsymbol{\epsilon}$,
$\boldsymbol{\epsilon} \sim \mathcal{N}(\mathbf{0},
\mathbf{I})$, with noise scale $\eta$. To reach the budget $N_{\max}$, primitives are
added either indirectly through HYDRA decomposition, or by
sampling additional primitives for relocation. In all cases
$P(\mathbf{g}^{\mathrm{new}}) \approx
P(\mathbf{g}^{\mathrm{old}})$ is maintained.

{\bf HYDRA: Hybrid Decomposition for Rendering-Aware Preservation.}
Under-reconstructed regions accumulate high positional gradients,
as observed in~\cite{kerbl_3d_2023,ye_absgs_2024}. Since we do
not apply positional SGLD noise, these
regions risk being undersampled across densification steps. The
most effective solution is to subdivide the primitives covering
such regions, increasing the number of independent samples that
can explore the local geometry. HYDRA is our mechanism for
performing this subdivision while preserving
$P(\mathbf{g}^{\mathrm{new}}) \approx
P(\mathbf{g}^{\mathrm{old}})$. We identify high-gradient
primitives via absolute gradient
accumulation~\cite{ye_absgs_2024} and distinguish two cases:

\emph{(i)~Scale-preserving split.} For smaller
primitives, we reduce the circumradius and place two children
symmetrically offset in the parent's tangent plane, applying the
mass-preserving relocation formulas of
\cref{eq:relocation_denom,eq:scale_relocation} with $N\!=\!2$.

\begin{figure*}[t]
  \centering
  \includegraphics[width=\linewidth]{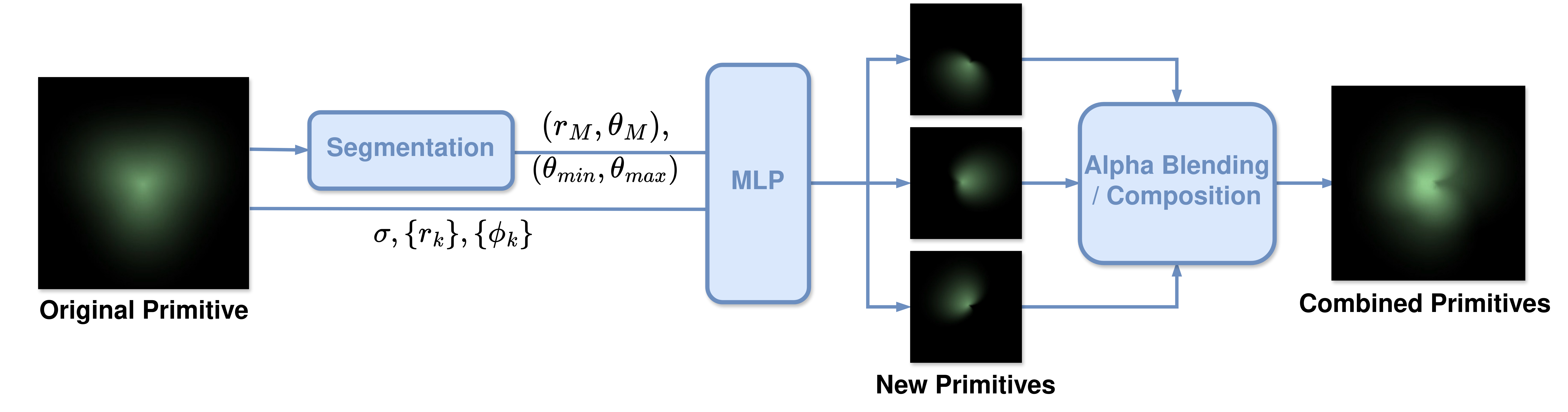}
  \caption{Learned lobe decomposition overview}
  \label{fig:mlp}
\end{figure*}

\noindent\emph{(ii)~Learned lobe decomposition.} For large
primitives that have developed complex, multi-lobed boundaries,
a geometric split is insufficient. We identify the $S$ deepest
valleys along the boundary $r_i(\theta)$ sampled at $M$ angles,
partitioning it into $S$ angular segments. A pre-trained MLP
$f_\psi$ generates a child primitive for each segment:
\begin{equation}
  f_\psi\!: \left(\sigma,\, \{\phi_k\},\, \{r_k\},\,
  r_M,\, \theta_M,\, \theta_{\min},\,
  \theta_{\max}\right) \mapsto \left(\sigma',\,
  \{\phi'_k\},\, \{r'_k\},\, \Delta r,\,
  \Delta\theta,\, o',\, R'\right),
  \label{eq:hydra}
\end{equation}
where $(r_M, \theta_M)$ and $(\theta_{min}, \theta_{max})$ denote the center of mass and angular bounds of the segment respectively and $(\Delta r, \Delta\theta)$ is the polar
offset used to compute the child center in the parent's tangent
plane. Children inherit color and orientation from the parent,
the parent is removed, yielding a net increase of $S-1$
primitives. \cref{fig:mlp}

 

{\bf Loss.}
We optimize a photometric objective combining $\ell_1$ and D-SSIM:
\begin{equation}
  \mathcal{L}_{\mathrm{photo}} = (1-\lambda)\,\mathcal{L}_1
  + \lambda\,\mathcal{L}_{\mathrm{D\text{-}SSIM}}.
  \label{eq:photo_loss}
\end{equation}
Following 2DGS~\cite{huang_2d_2024}, we add a depth distortion loss
$\mathcal{L}_{\mathrm{dist}}$ penalizing spread in rendered depth
distributions, and a normal consistency loss
$\mathcal{L}_{\mathrm{normal}}$ aligning each primitive's normal
$\mathbf{t}_w$ with the depth-derived surface normal:
\begin{equation}
  \mathcal{L} = \mathcal{L}_{\mathrm{photo}}
  + \lambda_{\mathrm{dist}}\,\mathcal{L}_{\mathrm{dist}}
  + \lambda_{\mathrm{normal}}\,\mathcal{L}_{\mathrm{normal}}.
  \label{eq:total_loss}
\end{equation}

\section{Experiments}
\label{sec:experiments}

We evaluate Fourier Splatting against state-of-the-art methods in novel view synthesis
across standard benchmarks, and then analyze the scalability properties that distinguish
our approach.
\subsection{Experimental Setup}
\label{sec:setup}


{\bf Datasets, metrics, and baselines.}
We evaluate on Mip-NeRF 360~\cite{barron_mip-nerf_2022} (5
outdoor, 4 indoor scenes) and Tanks and
Temples~\cite{knapitsch_tanks_2017} (2 large-scale scenes),
measuring PSNR, SSIM~\cite{wang_image_2004}, and
LPIPS~\cite{zhang_unreasonable_2018}. We compare against implicit
methods (Mip-NeRF 360~\cite{barron_mip-nerf_2022},
Zip-NeRF~\cite{barron_zip-nerf_2023}), volumetric primitives
(3DGS~\cite{kerbl_3d_2023},
3DGS-MCMC~\cite{kheradmand_3d_2024},
DBS~\cite{liu_deformable_2025},
3DCS~\cite{held_3d_2025},
Gabor Splatting~\cite{zhou_3dgabsplat_2025}, GES~\cite{hamdi_ges_2024}), and planar
primitives (2DGS~\cite{huang_2d_2024}, Triangle
Splatting~\cite{held_triangle_2025},
BBSplat~\cite{svitov_billboard_2025}).
We follow the evaluation protocol of Triangle Splatting~\cite{held_triangle_2025} and report their published
baseline numbers, additional results are taken from
DBS~\cite{liu_deformable_2025}.

{\bf Implementation details.}
We use $K\!=\!6$ Fourier frequencies and spherical harmonics of
degree three, yielding 70 parameters per primitive. The Fourier
boundary (\cref{eq:fourier_boundary}) is evaluated via Horner's
method. We compute $w = e^{\mathrm{i}\theta} = (\cos\theta,
\sin\theta)$ directly from the tangent-plane coordinates as
$\cos\theta = u_j / \rho_j$, $\sin\theta = v_j / \rho_j$,
avoiding all transcendental calls. The polynomial is then
accumulated as a recurrence from $k\!=\!K\!-\!1$ down to $0$:
\begin{equation}
  z_{K-1} = c_{i,K-1}, \qquad
  z_{k} = z_{k+1} \cdot w + c_{i,k},
  \label{eq:horner}
\end{equation}
yielding $r_i(\theta) = R_i |z_0|$ in $K\!-\!1$ complex
multiply-adds, without evaluating $e^{\mathrm{i}k\theta}$ for any
$k > 1$. The optimization process begins by activating only the lowest frequency components $K_{\mathrm{active}}\!=\!1$ to establish a stable global structure. At iteration 600, the remaining frequency coefficients are unfrozen, preventing low-frequency dominance from obstructing the acquisition of finer textural details. Due to the orthonormality of the Fourier basis, the optimal $L^2$ approximation of an $K$-term polynomial using $K-1$ coefficients is obtained by simple truncation. Because the expansion coefficients are independent, the remaining terms require no adjustment to minimize the residual energy.

\subsection{Novel View Synthesis}
\label{sec:nvs}

\begin{figure}[t]
  \centering
  \includegraphics[width=\linewidth]{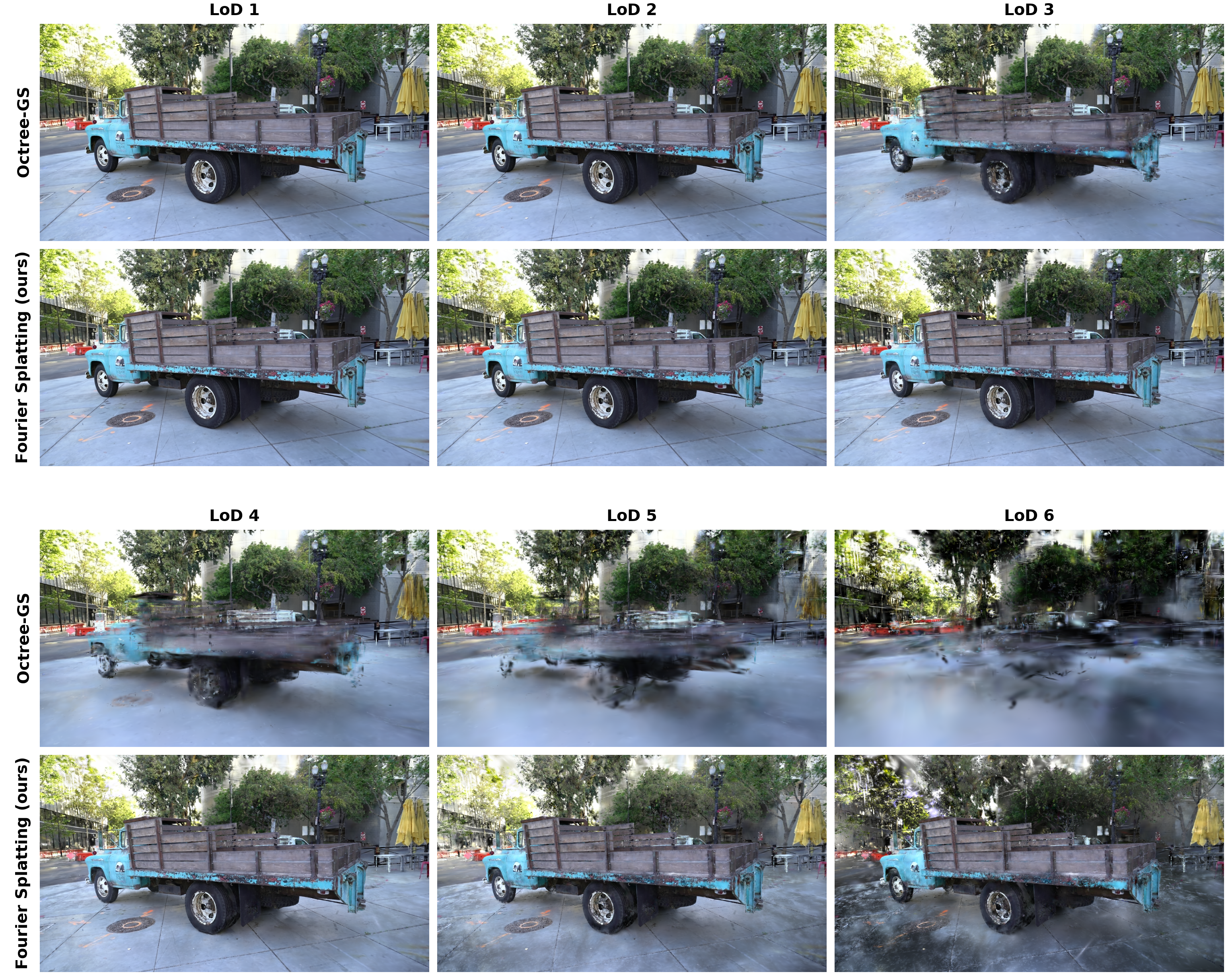}
  \caption{\textbf{Qualitative Scalability Analysis.} Comparison of the different LoDs between the proposed method and Octree-GS~\cite{ren_octree-gs_2025} on the Truck from Tanks and Temples.}
  \label{fig:scalability_qual}
\end{figure}

{\bf Quantitative results.}
\cref{tab:main_results} presents a comprehensive evaluation on the Mip-NeRF 360 and Tanks and
Temples datasets. Among methods using non-volumetric primitives, our approach achieves state-of-the-art performance, delivering the best PSNR and SSIM on both indoor and outdoor Mip-NeRF scenes. Specifically, outperforming Triangle Splatting~\cite{held_triangle_2025} by 0.67~dB on the Mip-NeRF 360 average.
Compared to 2DGS~\cite{huang_2d_2024}, which shares our planar surfel formulation, the
gains are more pronounced (+0.81~dB). These significant gains validate that the generalised boundary is able to achieve higher reconstruction quality. Furthermore, Fourier Splatting competes favorably against volumetric methods, we match or
exceed DBS~\cite{liu_deformable_2025} on average LPIPS despite our purely planar
primitive, and surpass 3DGS~\cite{kerbl_3d_2023} across all metrics. On Tanks
and Temples, our method achieves the top LPIPS score among all tested baselines while remaining highly competitive in PSNR and SSIM.

\begin{table*}[t]
\centering
\setlength{\tabcolsep}{3.5pt}
\renewcommand{\arraystretch}{1.05}
\caption{Quantitative comparison on Mip-NeRF
360~\cite{barron_mip-nerf_2022} and Tanks \&
Temples~\cite{knapitsch_tanks_2017}. Among planar primitives:
\colorbox{red!20}{\textbf{best}},
\colorbox{orange!20}{second best}. We follow the evaluation
protocol of Triangle Splatting~\cite{held_triangle_2025}. Our method ranks first
among planar methods on Mip-NeRF 360 across all metrics and
achieves the best SSIM and LPIPS on Tanks \& Temples.}
\label{tab:main_results}
\resizebox{\textwidth}{!}{%
\begin{tabular}{l ccc ccc ccc ccc}
\toprule
& \multicolumn{3}{c}{Outdoor Mip-NeRF 360}
& \multicolumn{3}{c}{Indoor Mip-NeRF 360}
& \multicolumn{3}{c}{Avg.\ Mip-NeRF 360}
& \multicolumn{3}{c}{Tanks \& Temples} \\
\cmidrule(lr){2-4} \cmidrule(lr){5-7} \cmidrule(lr){8-10} \cmidrule(lr){11-13}
Method
& PSNR$\uparrow$ & SSIM$\uparrow$ & LPIPS$\downarrow$
& PSNR$\uparrow$ & SSIM$\uparrow$ & LPIPS$\downarrow$
& PSNR$\uparrow$ & SSIM$\uparrow$ & LPIPS$\downarrow$
& PSNR$\uparrow$ & SSIM$\uparrow$ & LPIPS$\downarrow$ \\
\midrule
\multicolumn{13}{l}{\textit{Implicit Methods}} \\
Mip-NeRF 360~\cite{barron_mip-nerf_2022}
  & 24.47 & 0.691 & 0.283
  & 31.72 & 0.917 & 0.179
  & 27.35 & 0.792 & 0.237
  & 22.22 & 0.759 & 0.257 \\
Zip-NeRF~\cite{barron_zip-nerf_2023}
  & 25.58 & 0.750 & 0.207
  & 32.25 & 0.926 & 0.167
  & 28.42 & 0.826 & 0.189
  & -- & -- & -- \\
\midrule
\multicolumn{13}{l}{\textit{Volumetric Primitives}} \\
3DGS~\cite{kerbl_3d_2023}
  & 24.64 & 0.731 & 0.234
  & 30.41 & 0.920 & 0.189
  & 26.98 & 0.813 & 0.214
  & 23.14 & 0.841 & 0.183 \\
3DGS-MCMC$^{\ddagger}$~\cite{kheradmand_3d_2024}
  & 25.51 & 0.760 & 0.210
  & 31.08 & 0.917 & 0.208
  & 27.84 & 0.850 & 0.210
  & 24.29 & 0.860 & 0.190 \\
DBS$^{\dagger}$~\cite{liu_deformable_2025}
  & 25.10 & 0.751 & 0.246
  & 32.29 & 0.936 & 0.220
  & 28.13 & 0.827 & 0.234
  & 24.85 & 0.870 & 0.140 \\
GES~\cite{hamdi_ges_2024}
  & 24.89 & 0.717 & 0.288
  & 27.60 & 0.856 & 0.233
  & 26.91 & 0.794 & 0.250
  & 23.35 & 0.836 & 0.198 \\
3DCS~\cite{held_3d_2025}
  & 24.07 & 0.700 & 0.238
  & 31.33 & 0.927 & 0.166
  & 26.99 & 0.800 & 0.207
  & 23.94 & 0.851 & 0.156 \\
Octree-GS-Scaffold~\cite{ren_octree-gs_2025}
    & 24.83 & 0.724 & 0.262
    & 32.08 & 0.937 & 0.155 
    & 28.05 & 0.819 & 0.214
    & 24.68 &  0.866 & 0.153 \\
\midrule
\multicolumn{13}{l}{\textit{Planar Primitives}} \\
2DGS~\cite{huang_2d_2024}
  & \cellcolor{orange!20}24.34 & 0.717 & 0.246
  & 30.40 & 0.916 & 0.195
  & 26.84 & 0.804 & 0.252
  & 23.13 & 0.831 & 0.212 \\
BBSplat$^{\dagger}$~\cite{svitov_billboard_2025}
  & 23.55 & 0.669 & 0.281
  & 30.62 & 0.921 & 0.178
  & 26.49 & 0.778 & 0.236
  & \cellcolor{red!20}\textbf{25.12} & \cellcolor{red!20}\textbf{0.868} & \cellcolor{orange!20}0.172 \\
Tri.\ Splatting~\cite{held_triangle_2025}
  & 24.27 & \cellcolor{orange!20}0.722 & \cellcolor{red!20}\textbf{0.217}
  & \cellcolor{orange!20}30.80 & \cellcolor{orange!20}0.928 & \cellcolor{red!20}\textbf{0.160}
  & \cellcolor{orange!20}26.98 & \cellcolor{orange!20}0.812 & \cellcolor{red!20}\textbf{0.191}
  & 23.14 & \cellcolor{orange!20}0.857 & 0.143 \\
\rowcolor{blue!6}
\textbf{Ours}
  & \cellcolor{red!20}\textbf{24.62} & \cellcolor{red!20}\textbf{0.739} & \cellcolor{red!20}\textbf{0.217}
  & \cellcolor{red!20}\textbf{31.44} & \cellcolor{red!20}\textbf{0.930} & \cellcolor{orange!20}0.162
  & \cellcolor{red!20}\textbf{27.65} & \cellcolor{red!20}\textbf{0.824} & \cellcolor{orange!20}0.193
  & \cellcolor{orange!20}24.15 & \cellcolor{red!20}\textbf{0.868} & \cellcolor{red!20}\textbf{0.137} \\
\bottomrule
\end{tabular}}
\end{table*}

\begin{figure*}[t]
  \centering
  \includegraphics[width=\linewidth]{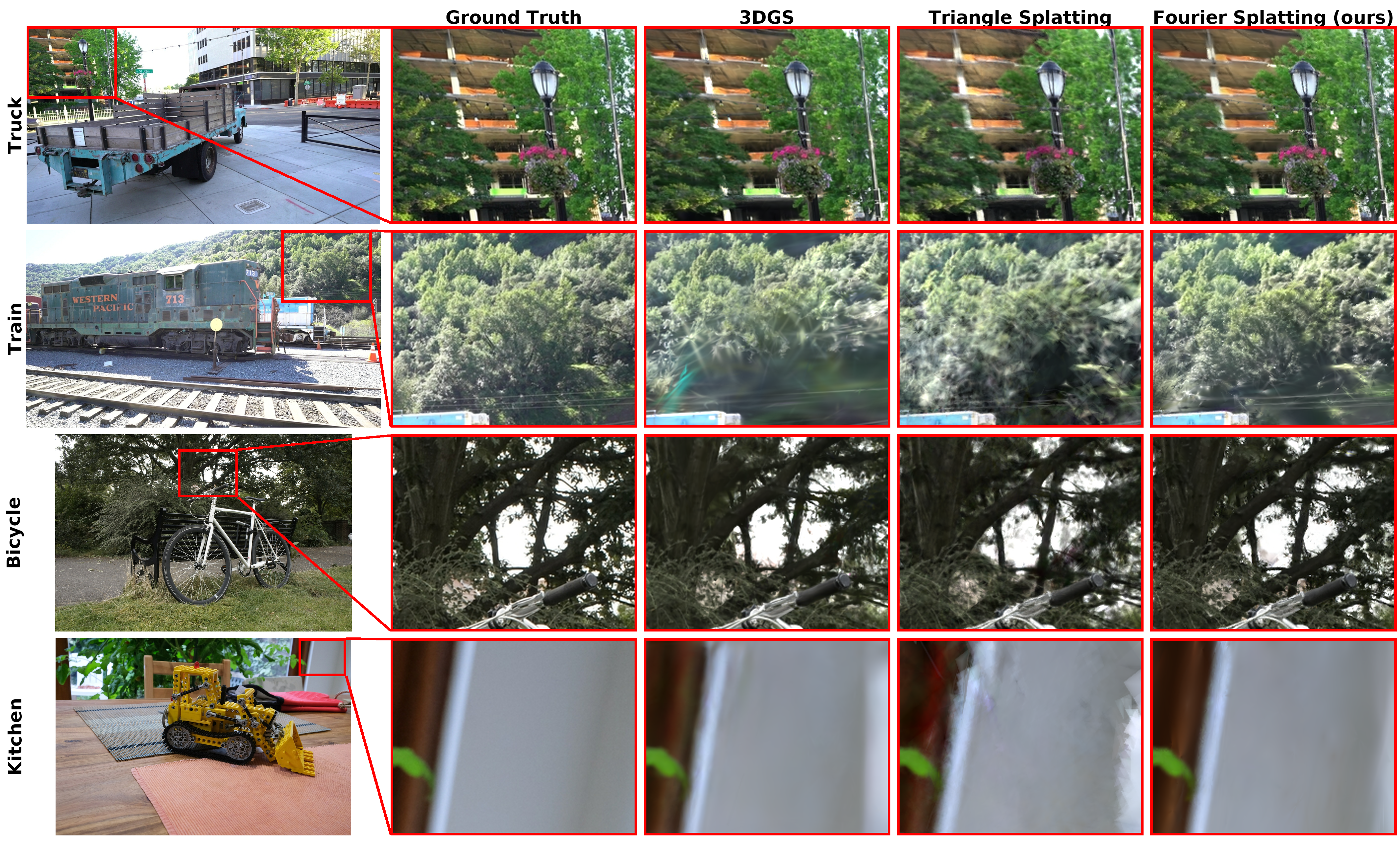}
  \caption{\textbf{Qualitative Analysis.} Comparison between the ground truth, state-of-the-art methods and our proposed approach on 2 scenes from Mip-NeRF 360 and Tanks and Temples.}
  \label{fig:qualitative}
\end{figure*}

{\bf Qualitative results.}
\cref{fig:qualitative} shows visual comparisons on representative scenes from Mip-NeRF 360 ~\cite{barron_mip-nerf_2022} and Tanks and Temples ~\cite{knapitsch_tanks_2017}. Fourier Splatting recovers fine details---thin structures, high-frequency textures, and sharp silhouettes---more faithfully than 3DGS, which tends to blur these regions due to the limited expressiveness of its geometry. Compared to Triangle Splatting, our method is less prone to artifacts where the underlying primitive shape is clearly visible. We refer to the supplementary material for per-scene breakdowns and additional visual comparisons.\looseness=-1

\subsection{Scalability Analysis}
\label{sec:scalability}


A key contribution of our Fourier Splatting approach is the ability to scale primitive expressiveness independently of primitive count. This scalability requires no additional training or computation; image quality is adjusted simply by modulating the number of coefficients transmitted, up to the maximum $K$ used during training. In \cref{fig:k_ablation}, we demonstrate the monotonic decrease in quality as coefficients are truncated. \cref{fig:rate_distortion} provides a quantitative rate--distortion comparison against surface based primitive level-of-detail representation ~\cite{ren_octree-gs_2025}, with \cref{fig:scalability_qual} showing the corresponding qualitative degradation. While current scalable splatting methods like Octree-GS degrade quality by reducing the primitive count, which can cause focal objects to vanish, our method scales the expressiveness of the primitives themselves. This results in a more graceful and uniform degradation of visual quality.


{\bf Compression.}
Another benefit of the per-primitive scalability is the compatibility with compression codecs, which often code attributes independently. In contrast, scaling by primitive count complicates efficient geometry compression. We demonstrate this benefit by compressing a scene using the ISO V3C codec, resulting in independent bitstreams per primitive, facilitating efficient scalable streaming (Tab.~\ref{table:compression}).

\begin{table}[h]
\centering
\footnotesize
\setlength{\tabcolsep}{3pt}
\renewcommand{\arraystretch}{1.0}
\caption{Quantitative results for compressed scalability with V3C.}
\begin{tabular}{lcccccc}
\toprule
K & 1 & 2 & 3 & 4 & 5 & 6 \\
\midrule
Bytes/primitive & 61.92 & 64.52 & 67.13 & 69.67 & 72.21 & 74.89 \\
PSNR  & 12.66 & 16.74 & 19.18 & 20.77 & 21.81 & 22.40 \\
SSIM  & 0.53  & 0.66  & 0.74  & 0.79  & 0.82  & 0.84  \\
LPIPS & 0.45  & 0.35  & 0.27  & 0.22  & 0.19  & 0.17  \\
\bottomrule
\end{tabular}
\label{table:compression}
\end{table}

\subsection{Ablation Studies}
\label{sec:ablation}

We ablate the key components of Fourier Splatting on Tanks and
Temples (\cref{tab:ablation}). Removing {\bf automatic lobe
decomposition} degrades PSNR by 0.22~dB, as naive splitting
cannot decompose multi-lobed primitives into simpler constituents
that better fit the local geometry. Disabling the {\bf straight-through estimator} causes the largest drop of 0.34~dB:
without synthetic gradients outside the boundary, primitives
cannot grow to cover under-reconstructed regions and the
optimization stalls near circular shapes. By removing the {\bf
randomized cloning offset} the representation loses 0.05~dB
in PSNR, as the stochastic angular placement of added primitives in the surfel plane encourages spatial diversity.
\section{Limitations and Future Work}
\label{sec:limitations}

\begin{figure*}[t]
\centering

\begin{minipage}[b]{0.53\textwidth}
  \centering
  \setlength{\tabcolsep}{4pt}
  \renewcommand{\arraystretch}{1.05}
  {\small\captionof{table}{Comparison at matched 960K primitive budget (T\&T).}\label{tab:rebuttal}}
  \resizebox{\linewidth}{!}{%
  \begin{tabular}{l ccccccc}
  \toprule
  Method & $N_{\mathrm{prim}}$ & PSNR$\uparrow$ & SSIM$\uparrow$ & LPIPS$\downarrow$ & FPS & VRAM & Train \\
  \midrule
  2DGS           & 960K & 22.94          & 0.827          & 0.160          & \textbf{182} & \textbf{1.4\,GB} & \textbf{12\,min} \\
  Ours ($K{=}6$) & 960K & \textbf{23.79} & \textbf{0.853} & \textbf{0.156} & 135          & 2.08\,GB         & 41\,min \\
  \bottomrule
  \end{tabular}}
\end{minipage}\hfill
\begin{minipage}[b]{0.44\textwidth}
  \centering
  \setlength{\tabcolsep}{5pt}
  \renewcommand{\arraystretch}{1.05}
  {\small\captionof{table}{Ablation on Tanks and Temples.}\label{tab:ablation}}
  \resizebox{\linewidth}{!}{%
  \begin{tabular}{lccc}
  \toprule
  Configuration & PSNR$\uparrow$ & SSIM$\uparrow$ & LPIPS$\downarrow$ \\
  \midrule
  Full model              & \textbf{24.15} & \textbf{0.868} & \textbf{0.137} \\
  w/o lobe decomposition  & 23.93          & 0.847          & 0.141 \\
  w/o STE                 & 23.81          & 0.842          & 0.147 \\
  w/o cloning noise       & 24.10          & 0.845          & 0.138 \\
  \bottomrule
  \end{tabular}}
\end{minipage}

\vspace{0.6em}

\begin{minipage}[b]{0.48\textwidth}
  \centering
  \includegraphics[width=\linewidth]{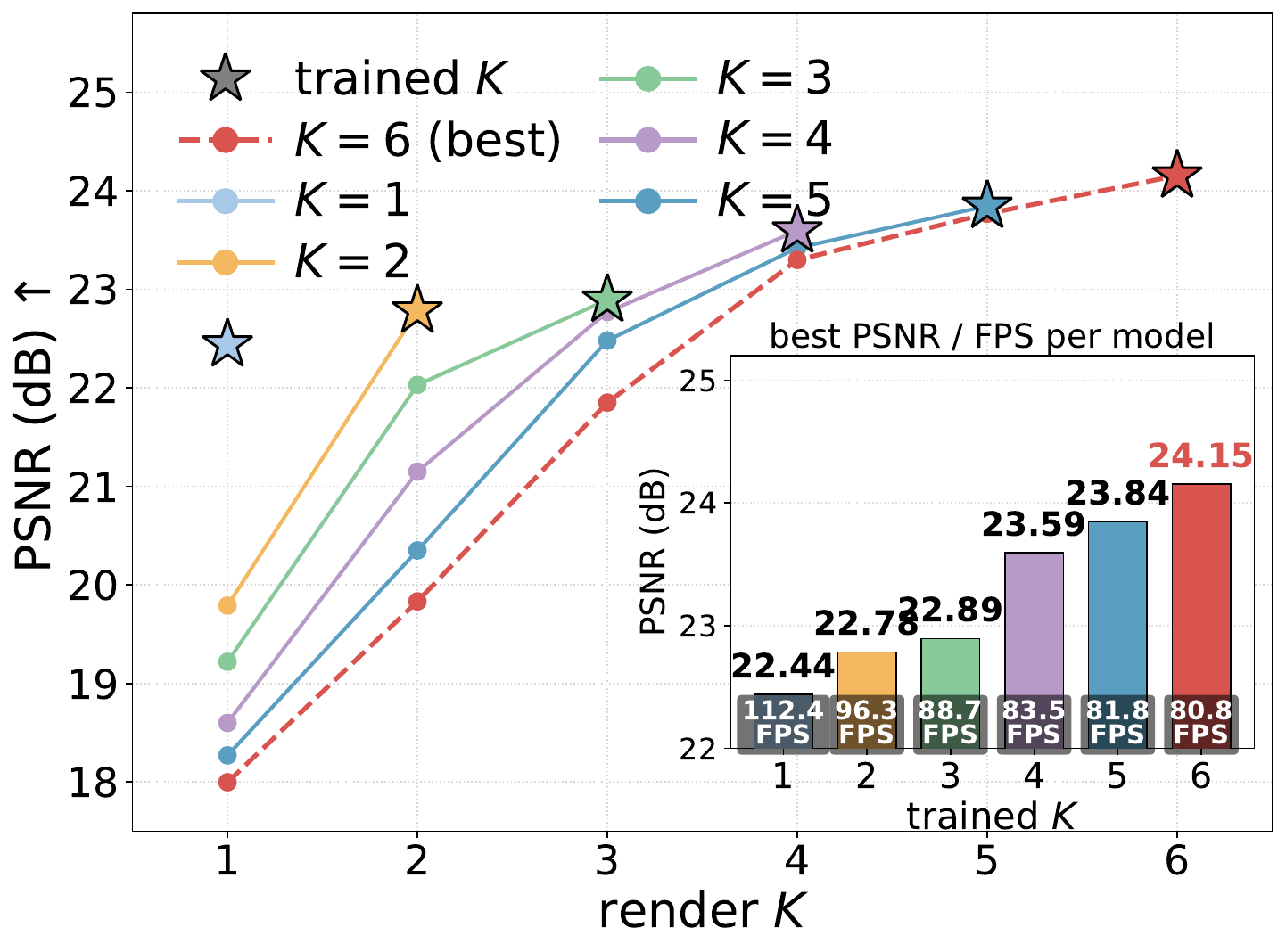}
  {\small\captionof{figure}{Ablation on trained $K$.}\label{fig:k_ablation}}
\end{minipage}\hfill
\begin{minipage}[b]{0.48\textwidth}
  \centering
  \includegraphics[width=\linewidth]{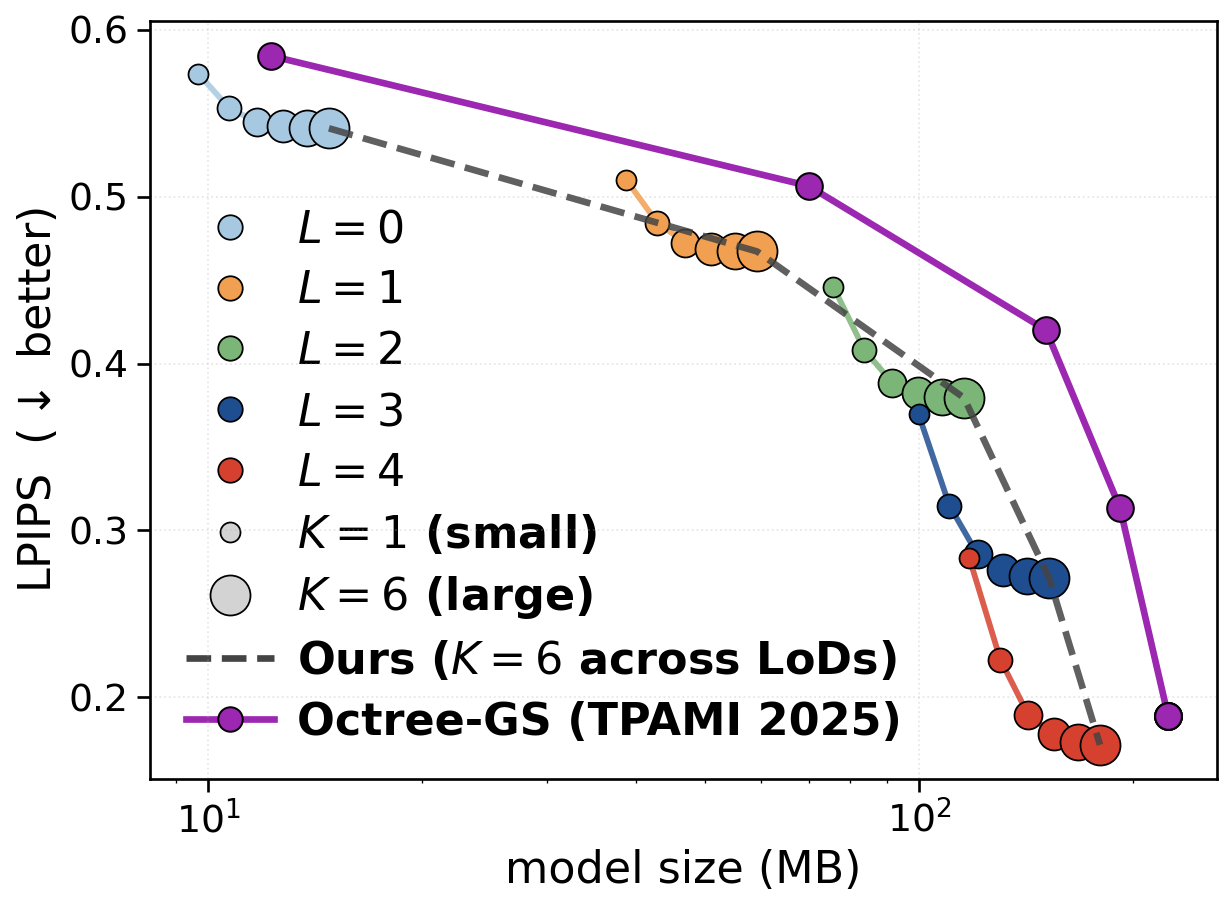}
  {\small\captionof{figure}{Rate--distortion comparison.}\label{fig:rate_distortion}}
\end{minipage}

\end{figure*}

{\bf Bitrate.}
The proposed radiance field representation prioritizes achieving
state-of-the-art fidelity for a given primitive budget rather than minimizing bitrate required to compress and transmit such representations. A promising avenue for future research involves the integration of scale regularization to incentivize the formation of fewer, more spatially expansive, and complex primitives. By leveraging the high-frequency expressiveness of high-$K$ boundaries to represent regions typically requiring dense distributions of simple surfels, such an approach could yield significant improvements in compression efficiency and overall bitrate reduction.

\noindent {\bf Computational Complexity.}
Despite their richer geometry, our primitives remain real-time: \cref{tab:rebuttal} reports 135 FPS at a matched 960K budget. The Horner recurrence keeps the overhead small, adding only 8 operations per frequency.

\section{Conclusion}
\label{sec:conclusion}


Fourier Splatting introduces a novel geometric primitive capable of approximating arbitrary planar shapes, marking the first representation for real-time radiance field rendering with inherently scalable expressiveness. By parameterizing surfel boundaries via a truncatable Fourier encoded descriptors, our method effectively decouples reconstruction fidelity from primitive density. This allows dynamic downscaling of visual quality and bitrate via coefficient truncation rather than primitive removal.
To enable stable optimization of these expressive primitives within a fixed budget, we employ a straight-through estimator that extends gradients beyond the primitive boundary and introduce HYDRA, a learned decomposition strategy that splits complex primitives into simpler constituents within the MCMC densification framework. Extensive evaluations on Mip-NeRF 360 and Tanks and Temples demonstrate that Fourier Splatting achieves state-of-the-art rendering quality among surface-based methods and remains competitive with volumetric approaches. Furthermore, our scalability analysis underscores the potential of per-primitive refinement as a robust alternative to traditional pruning. When integrated with classic pruning techniques, our approach provides a versatile framework for high-fidelity rendering in bandwidth-constrained environments.

\subsubsection{\bf Acknowledgments}  \emph{} \\ The work is supported by FWO, project Panoptes (G0ACE26N) and BELSPO, project ATTENTION.
 

%
\bibliographystyle{splncs04}
\bibliography{references}
\end{document}